\documentclass[a4paper]{article}

\usepackage{ISCSLP2022}
\usepackage{diagbox}
\usepackage{cite}
\usepackage{algorithm}
\usepackage{CJKutf8}
\usepackage{url}
\usepackage{CJK}
\usepackage{float}
\usepackage{algpseudocode}
\usepackage{multirow}
\usepackage{arydshln}
\usepackage{amsmath}

\title{Multi-Level Modeling Units for End-to-End Mandarin Speech Recognition}

\name{Yuting Yang, Binbin Du, Yuke Li$^*$\thanks{$^*$ Corresponding author}}
\address{NetEase Yidun AI Lab, Hangzhou, China}
\email{$\left\{yangyuting04, dubinbin, liyuke\right\}$@corp.netease.com}

\begin{document}

\maketitle
\begin{abstract}
The choice of modeling units is crucial for automatic speech recognition (ASR) tasks. In mandarin scenarios, the Chinese characters represent meaning but are not directly related to the pronunciation. Thus only considering the writing of Chinese characters as modeling units is insufficient to capture speech features. In this paper, we present a novel method involves with multi-level modeling units, which integrates multi-level information for mandarin speech recognition. Specifically, the encoder block considers syllables as modeling units and the decoder block deals with character-level modeling units. To facilitate the incremental conversion from syllable features to character features, we design an auxiliary task that applies cross-entropy (CE) loss to intermediate decoder layers. During inference, the input feature sequences are converted into syllable sequences by the encoder block and then converted into Chinese characters by the decoder block. Experiments on the widely used AISHELL-1 \cite{bu2017aishell} corpus demonstrate that our method achieves promising results with CER of 4.1\%/4.6\% and 4.6\%/5.2\%, using the Conformer and the Transformer backbones respectively. 
\end{abstract}

\section{Introduction}
Speech recognition is a technology to translate human speech into textual representation. It is of importance for human-computer interaction, and is widely used in various scenarios. Conventional speech recognition systems consist of three separate components, including an acoustic model, a pronunciation model as well as a language model. The acoustic model calculates the probability distribution of each frame over the modeling units. The pronunciation model maps the modeling units, e.g., context-dependent states (CD-states) or context-dependent phonemes (CD-phonemes), into words. The language model scores different word combinations. These three components are trained separately. In the inference phase, the conventional ASR systems usually compose the phonetic model and the language model to build a decoding graph using a weighted finite state transducer (WFST) \cite{mohri2002weighted} and finally obtain the decoding results through the Viterbi algorithm \cite{forney1973viterbi}. In recent years, with the development of deep learning, end-to-end (E2E) ASR systems have become a hot spot both for industrial landing and academic research. End-to-end speech recognition systems merge the acoustic model, the pronunciation model, and the language model into a big neural network. Compared with the traditional speech recognition system, end-to-end ASR systems have the advantages of simpler architecture and better performance.

The choice of modeling units is related to the inherent characteristics of the language. English words usually consist of several pronunciation units, while Chinese is a different scenario. Chinese characters are text symbols that are difficult to associate with pronunciation. Therefore, modeling with Chinese characters does not introduce phonological information into the network. On the other hand, Chinese has some unique characteristics, such as the pronunciation is toned and each Chinese character can be represented by a toned syllable (Pinyin with tone), which can be considered as pronunciation-related modeling units. 

In this paper, we study how to organically combine both the syllable and the character modeling units in end-to-end ASR models. We hypothesize that both the phonological and the context-based information are important for system performance. To achieve this, we propose a novel method involves with multi-level modeling units, in which the encoder block considers syllables as modeling units and the decoder block deals with the character modeling units. The encoder block introduces the phonological information into the model by learning the alignment between the feature sequences and the syllable sequences, and the decoder block transducers syllable sequences into output characters. During inference, the input feature sequences are converted into syllable sequences by the encoder block and then converted into Chinese characters by the decoder block in a non-autoregressive way.

In this work, the inference is a two-pass process conducted by a unified end-to-end model. Specifically, the encoder module plays the role of converting feature sequences to syllable sequences. And the decoder plays the role of converting syllables to Chinese characters, which is a more challenging task compared with the character-to-character decoder module. For the above considerations, we introduce an auxiliary task termed \textit{InterCE} to facilitate the conversion from syllable sequences to character sequences and improve the performance.

Our contributions are summarized as follows:
\begin{itemize}
    \item We propose a novel end-to-end ASR model equipped with both the syllable-level and the character-level modeling units, which utilizes phonological as well as context-based information during the training process.
    \item We introduce an auxiliary task \textit{InterCE} into the decoder block to further facilitate the conversion from syllable sequences to character sequences.
    \item We apply multi-level modeling units to Transformer \cite{vaswani2017attention} and Conformer \cite{gulati2020conformer} architectures respectively, and show the proposed method is effective for both architectures.
\end{itemize}

\section{Related Work}
\subsection{E2E ASR}
In recent years, end-to-end speech recognition (E2E ASR) systems have achieved outstanding improvements and become more and more popular. One research line is the connectionist temporal classification (CTC) \cite{2006Connectionist} based methods. CTC-based ASR models calculate the probability distribution of each frame over the dictionary to obtain the alignment between the input and output sequences. In addition, recurrent neural network transducer (RNN-T) \cite{graves2012sequence} is very popular in streaming on-device speech recognition since its low-latency and promising performance. 

Another cluster is the attention-based models \cite{chan2016listen,2015Attention,vaswani2017attention}. Transformer \cite{vaswani2017attention} is a dominant model among them, which has achieved state-of-the-art performance on neural machine translation (NMT). It adopts the encoder-decoder structure, both of which consist of several stacked sub-layers. Self-attention is the foundation module of Transformer, which is a mechanism linking all the position-pairs of a sequence to capture content-based global interactions. Transformer is also applicable to the task of speech recognition, in which the typical model is Speech-Transformer \cite{dong2018speech}. Speech-Transformer has been widely used in the field of ASR due to the nature of parallel training, the ability to capture long distance contextual information and its outstanding performance. Many efforts have been done on modifying the architecture of the Transformer to achieve better performance on speech recognition \cite{mohamed2019transformers,gulati2020conformer}. For instance, Conformer \cite{gulati2020conformer} is a convolution-augmented Transformer, which combines Transformer with convolutional neural networks to extract both global and local dependencies of feature sequences. 

\subsection{E2E ASR with Auxiliary Tasks}
There are lots of works that explored the improvements of auxiliary tasks for ASR based on the assumption that different layers learn representations at different levels \cite{krishna2018hierarchical,sanabria2018hierarchical,toshniwal2017multitask,tjandra2020deja,lee2021intermediate,nozaki2021relaxing}. 
Interlayer CTC \cite{lee2021intermediate} adopts the auxiliary CTC loss on the CTC-based model, it uses the intermediate representation of the encoder layer to calculate the auxiliary CTC loss. Auxiliary tasks are applied to the Transformer \cite{tjandra2020deja} and RNN-T networks \cite{liu2021improving} as well. In this work, We apply an auxiliary task to the decoder block to further enhance the ability of converting syllable representations to character representations. We study the improvements of different types of InterCE tasks under the multi-level modeling units framework in Section 4.

\subsection{Modeling Units for Mandarin ASR}
Many works have explored how to choose suitable modeling units for different end-to-end systems in mandarin speech recognition \cite{zhang2019investigation,zhou2018comparison,zhou2018syllable,zou2018comparable}. 
In \cite{zou2018comparable}, three modeling units achieve similar CER but syllable modeling units are slightly superior to character modeling units for the LSTM-CTC model, and the character-level modeling units are better than other options for attention-based models. In \cite{zhou2018comparison}, five modeling units are compared based on Transformer structure, it draws conclusions that lexicon-free modeling units outperform lexicon-related modeling units and character is superior to other modeling units with Transformer models. 
All the above mentioned works show that character modeling units demonstrate good results for attention-based systems.

Many efforts have been made to exploit the pronunciation-related syllable modeling units for end-to-end systems, like the RNNT model \cite{wang2021cascade} and the attention-based model \cite{yuan2021decoupling}. They decouple the mandarin ASR task into two sub-tasks, including a recognition task converting the audio to syllables and a transcription task converting syllables to characters.
In these cases, it is necessary to introduce an additional model to convert modeling units into Chinese characters, which makes the system more complicated and introduces cumulative errors into the system. In addition, syllables also served as modeling units in the Mandarin-English code-switching ASR task \cite{chuang2021non}.

In this paper, we propose a novel end-to-end mandarin speech recognition method equipped with multi-level modeling units, in which the pronunciation-related information is utilized by the model while learning the alignment between the feature sequences and the syllable sequences. Moreover, the inference process is conducted by a unified end-to-end model without introducing any additional models.

\begin{figure}[t]
  \centering
  \includegraphics[width=\linewidth]{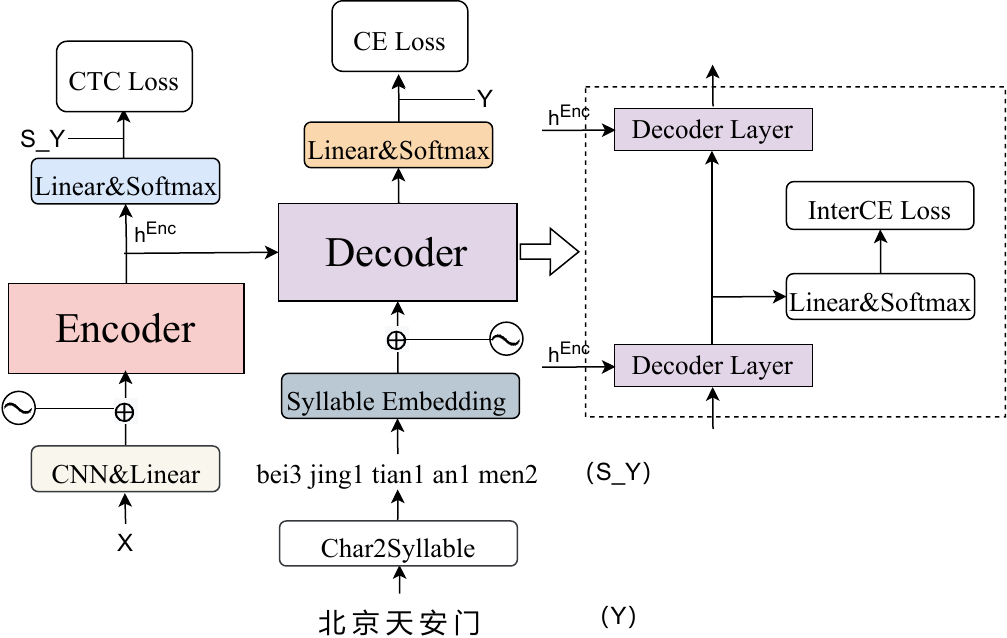}
  \caption{The framework with multi-level modeling units. ($X$, $S\_Y$, $Y$) denote the input features, the syllable sequence and the character sequence respectively. We take InterCE loss for the output of some intermediate decoder layers. The InterCE auxiliary tasks only exist in some special decoder layers and are used only in the training process.}
  \label{fig:speech_production}
\end{figure}

\section{Proposed Method}
We adopt two-level modeling units for the acoustic model, including character-level modeling units and syllable-level modeling units. We introduce the syllable-level modeling unit into the model to encode phonological information. At the same time, the character-level modeling units are retained to maintain the target character accuracy.

In the training process, we use the sequence of speech features ${\rm X}=\left\{x_1,x_2,x_3,...,x_T\right\}$ and the text sequence ${\rm Y}=\left\{y_1,y_2,y_3,...,y_n\right\}$ to train the network parameters. In this work, the syllable denotes Pinyin with tone. The syllable sequence ${\rm S\_Y}=\left\{sy_1,sy_2,sy_3,...,sy_n\right\}$ is obtained by an open source Chinese character to pinyin tool python-pinyin\footnote{\url{https://github.com/mozillazg/python-pinyin}}. The given input $\rm X$ is a feature matrix with length $T$ and dimension $D$. Each element $y_i \in {V_c}$ is a token of the character sequence $\rm Y$ and each element $sy_i \in {V_s}$ is a token of the syllable sequence $\rm S\_Y$, where $V_c$ is the character-level dictionary and $V_s$ is the syllable-level dictionary.
In the inference phase, we can obtain the text sequence $\rm Y$ given the speech feature sequence $\rm X$. 

\subsection{Model Architecture}
Figure~\ref{fig:speech_production} shows the framework with multi-level modeling units. We utilize a hybrid CTC/Attention method \cite{watanabe2017hybrid}, in which the CTC classifier considers syllables as modeling units, and the decoder deals with character modeling units. We adopt the encoder block with Transformer as well as Conformer backbones.
An encoder layer of Transformer is composed of two modules stacked together, i.e., a self-attention module and a feed-forward module. In contrast, an encoder layer of Conformer is composed of four modules, i.e., a feed-forward module, a self-attention module, a convolution module, and a second feed-forward module.
The decoder layer is composed of three modules stacked together, i.e., a masked self-attention module, a source-target multi-head attention module, and a feed-forward module. The encoder block consists of a stack of $N$ identical layers and the decoder block consists of stacked $L$ layers. 

\begin{table}[t]
  \caption{Character error rates with different $\alpha$ and $\beta$ values.
  }
  \label{tab:hyper-parameter}
  \centering
  \begin{tabular}{p{4cm}cc}
   \hline
    The value of ($\alpha$, $\beta$) & Dev set & Test set  \\
   \hline
    ($\alpha$=0.1, $\beta$=0.2) &4.2&4.6  \\
    ($\alpha$=0.1, $\beta$=0.4) &4.1&4.6 \\   
    ($\alpha$=0.3, $\beta$=0.2) &4.4&5.0 \\ 
   ($\alpha$=0.3, $\beta$=0.4) &4.4&4.9 \\ 
   \hline
  \end{tabular}
\end{table}

\begin{table*}[t]
  \caption{The descriptions of multi-level dictionaries.}
  \label{tab:dict}
  \centering
  \begin{tabular}{lll}
  \hline
	Modeling units&{Detailed composition} & {Examples} \\
	\hline
	\textit{Character-level} & 4233 tokens including unknown token $\langle$unk$\rangle$ and start/end token $\langle$sos/eos$\rangle$& \begin{CJK*}{UTF8}{gbsn}你\ 好\end{CJK*} \\
    \textit{Syllable-level} & 1370 tokens including $\langle$blank$\rangle$ token, $\langle$unk$\rangle$ token and $\langle$sos/eos$\rangle$ token&ni3 hao3\\
  \hline
  \end{tabular}
\end{table*}

The main difference between the NMT Transformer \cite{vaswani2017attention} and the ASR Transformer \cite{dong2018speech} is the input of the encoder. The ASR Transformer utilizes several convolutional layers to produce the hidden representations, and a subsequent linear layer to map the dimension of flattened features to the attention dimension $\rm d_{m}$. The outputs of the linear layer added with position encoding are fed to the encoder block. The encoder block extracts acoustic features represented $\rm h^{Enc}$ as
\begin{equation}
  {\rm h^{Conv} = Linear(Conv(X))},
  \label{eq-conv}
\end{equation}
\begin{equation}
  {\rm h^{Enc} = Encoder(h^{Conv})},
  \label{eq1}
\end{equation}
\begin{equation}
  {\rm z^{Enc} = Softmax(Linear_{d_{m}\rightarrow{|V_s|}}({h^{Enc}}))},
  \label{eq2}
\end{equation}
\begin{equation}
 {\rm \pounds_{ctc} = -log\ \textit{P}_{ctc} (S\_Y|z^{Enc}) },
  \label{eq3}
\end{equation}
where the outputs of the encoder block are applied to calculate syllable-level CTC loss. As shown in Eq.(\ref{eq2}), the outputs of the encoder block are fed into a linear layer to map the dimension of $\rm h^{Enc}$ to syllable-level dictionary dimension $\rm |V_s|$, $\rm Softmax(\cdot)$ means a softmax function to get the posterior probability distribution over the syllable-level dictionary. As shown in Eq.(\ref{eq3}), the outputs $\rm z^{Enc}$ together with the syllable labels $\rm S\_Y$ are applied to get the CTC loss $\rm \pounds_{ctc}$. 

Given the acoustic feature representation $\rm h^{Enc}$, the hidden states of hybrid modeling units can be formulated as
\begin{equation}
  {\rm  h^{Dec}_L = Decoder(h^{Enc},\ h^{Dec}_0)},
  \label{eq4}
\end{equation}
where $\rm h^{Dec}_0$ represents the $embedding$ of the syllable sequence $\rm S\_Y$ added with position encoding. The decoder uses the outputs of the encoder block and $\rm h^{Dec}_0$ to get the hidden representation $\rm h^{Dec}_L$, where $L$ is the number of decoder layers. According to this, the character-level CE loss is
\begin{equation}
  {\rm  c^{Dec}_L = Softmax(Linear_{d_{m}\rightarrow{|V_c|}}(h^{Dec}_L))},
  \label{eq5}
\end{equation}
\begin{equation}
  {\rm \pounds_{att} = -log\ \textit{P}_{att}(Y|c^{Dec}_L) }.
  \label{eq6} 
\end{equation}
where the outputs of the decoder block are fed into a linear layer to map the dimension from $d_{m}$ to $\rm |V_c|$ that is the size of the character-level dictionary. The final output $\rm c^{Dec}_L$ together with the character label $\rm Y$ are applied to get the CE loss $\rm \pounds_{att}$.

\begin{figure}[!t]
  \centering
  \includegraphics[width=\linewidth]{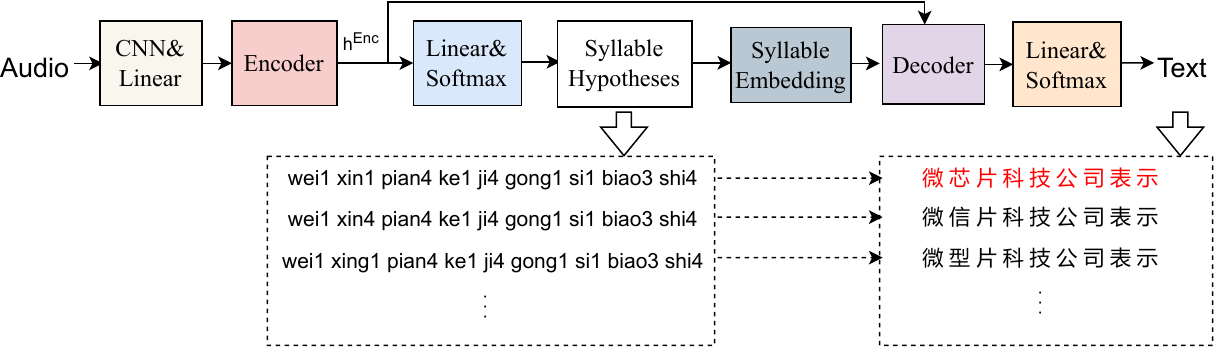}
  \caption{The pipeline of inference process.}
  \label{fig:inference}
\end{figure}


\subsubsection{\textit{InterCE} Loss}
We introduce interlayer auxiliary tasks to further facilitate the conversion from syllable sequences to character sequences. We call it \textit{InterCE} loss here. The schematic diagram of the \textit{InterCE} is provided in Figure~\ref{fig:speech_production}. We conduct the \textit{InterCE} loss using the outputs of some special layers of the decoder block. Formally,
\begin{equation}
  {\rm c^{Dec}_\epsilon = Softmax({\rm Linear}_{d_{m}\rightarrow{|V|}}(h^{Dec}_{\epsilon}))},
  \label{eq7}
\end{equation}
\begin{equation}
  {\rm \pounds_{\epsilon} = - log\ \textit{P}_{att}(Y'|c^{Dec}_{\epsilon})},
  \label{eq8}
\end{equation}
where $\epsilon$ is a selected interlayer of decoder block. $\rm (|V|,Y')=(|V_c|,Y)$ when use the character-level auxiliary loss, and $\rm (|V|,Y')=(|V_s|,S\_Y)$ with the syllable-level auxiliary loss. 

\begin{equation}
  {\rm \pounds = {\alpha} \pounds_{ctc} + {\beta}\frac{1}{k}\sum_{\epsilon \in{\mathcal{R}}} \pounds_{\epsilon} + (1-{\alpha}-{\beta}) \pounds_{att}}
  \label{eq9}
\end{equation}
When adding with \textit{InterCE} loss in the model, the total loss function is a linear combination among the syllable-level CTC loss, the \textit{InterCE} loss and the character-level CE loss.  As shown in Eq.(\ref{eq9}), where $\mathcal{R}$ denotes the set of intermediate decoder layers, and $k$ is the number of intermediate layers.

\section{Experiments}
We conduct experiments on AISHELL-1 \cite{bu2017aishell} which is a benchmark dataset for mandarin ASR, containing 178 hours collected from 400 people. We apply speed perturbation \cite{ko2015audio} and SpecAugment \cite{2019SpecAugment} for training data. For all experiments, the speech features are 80-dimensional log-mel filterbank (FBank) computed on 25ms window with 10ms shift. 
Table~\ref{tab:dict} describes the detailed composition
of two-level modeling units in this study, the character dictionary consists of 4233 tokens and the syllable dictionary consists of 1370 tokens.

\begin{table}[t]
  \caption{Comparison of different models: 
  SUER represents the syllable units error rate and CER represents the final character error rate. (I1,I2,I3,I4) denote different InterCE type.}
  \label{tab:transformer}
  \centering
  \begin{tabular}{lcccc}
  \hline
    \multirow{2}*{Methods}&\multicolumn{2}{c}{Dev set }&\multicolumn{2}{c}{Test set} \\
	&SUER&CER&SUER&CER \\
	\hline
	\multicolumn{5}{l}{\textbf{Character-based}} \\
     \hspace{0.2cm} RNN-T \cite{wang2021cascade}&-&-&-&6.1\\
     \hspace{0.2cm}Transformer\cite{yao2021wenet}&-&4.8&-&5.3 \\
	\hline
	\multicolumn{5}{l}{\textbf{Syllable-based}} \\
	\hspace{0.2cm}Cascade-RNNT\cite{wang2021cascade}&-&-&-&5.7 \\
	\hspace{0.2cm}Decouping+T \cite{yuan2021decoupling}&-&-&4.3&6.1\\
	\hspace{0.2cm}Decouping-T\cite{yuan2021decoupling}&-&-&2.8&7.3\\
    \hline 
    \multicolumn{5}{l}{\textbf{Our work}} \\
    \hspace{0.2cm}Transformer + I1&2.7&4.7&3.3&5.3\\
    \hspace{0.2cm}Transformer + I2&2.9&4.8&3.4&5.3\\
    \hspace{0.2cm}\textbf{Transformer + I3}&2.7&\textbf{4.6}&3.2&\textbf{5.2}\\
    \hspace{0.2cm}Transformer + I4&2.9&4.8&3.4&5.2\\
  \hline
  \end{tabular}
\end{table}

\begin{table}[t]
  \caption{Performance with Conformer model. SUER represents the syllable units error rate and CER represents the final character error rate. (I1,I2,I3,I4) denote different InterCE type.}
  \label{tab:conformer}
  \centering
    \begin{tabular}{lcccc}
  \hline
    \multirow{2}*{Methods}&\multicolumn{2}{c}{Dev set }&\multicolumn{2}{c}{Test set} \\
	&SUER&CER&SUER&CER \\
	\hline
	\multicolumn{5}{l}{\textbf{Character-based}} \\
	\hspace{0.2cm}Conformer + MMI \cite{tian2021consistent} &-&4.5&-&4.9\\
	\hspace{0.2cm}Conformer&-&4.4&-&4.7\\
	\hline
	\multicolumn{5}{l}{\textbf{Our work}} \\
    \hspace{0.2cm}Conformer + I1&2.2&4.1&2.6&4.6\\
    \hspace{0.2cm}Conformer + I2&2.2&4.1&2.6&4.6 \\
    \hspace{0.2cm}Conformer + I3&2.2&4.1&2.6&4.7\\
    \hspace{0.2cm}\textbf{Conformer + I4}&2.1&\textbf{4.1}&2.5&\textbf{4.6}\\
  \hline
  \end{tabular}
\end{table}
\subsection{Experimental Setup}
We follow the basic architectures of both Transformer~\cite{vaswani2017attention} and Conformer~\cite{gulati2020conformer}. The encoder is a 12-layer structure. The width of the attention layer is 256, the number of attention heads is 4, and the dimension in the feed-forward network sub-layer is 2048. The decoder is a 6-layer structure. Specifically, the kernel size is 15 for Conformer. The experiments are conducted using Wenet \cite{yao2021wenet}. The Adam optimizer \cite{kingma2014adam} with gradient clipping at 5.0 and warmup with 25000 steps are used during the training process. 
We employ dropout \cite{srivastava2014dropout} to prevent over-fitting.

We conduct experiments to search for the best hyper-parameters of $(\alpha, \beta)$. The results are shown in Table~\ref{tab:hyper-parameter}. As for $\alpha$, we found that the performance is greatly improved with the decrease of $\alpha$. This might be because converting syllables to characters is a challenging task. As the $\alpha$ decreases, the model pays more attention to enhancing conversion capabilities. As for $\beta$, the increase of $\beta$ has little effect on performance improvement. We set $\alpha=0.1$, $\beta=0.4$ in the following experiments.

We applied \textit{InterCE} loss to different decoder layers. We use $\mathcal{R}=\{3\}$ as \textit{default} layer and $\mathcal{R}=\{2,4\}$ as \textit{Multiple} layers. The different types of InterCE tasks are described as follows:
\begin{itemize}
   \item  \textbf{I1}: Multiple layers with syllable-level \textit{InterCE} loss.
   \item  \textbf{I2}: Multiple layers with character-level \textit{InterCE} loss.
   \item  \textbf{I3}: Default layer with syllable-level \textit{InterCE} losss.
   \item  \textbf{I4}: Default layer with character-level \textit{InterCE} loss. 
\end{itemize}

\subsection{Inference}
The pipeline of the inference process is shown in Figure~\ref{fig:inference}. In the inference process, the encoder module extracts acoustic features using the input feature sequence $X$. The encoder module is followed by a linear projection layer and a softmax function to get the probability distribution over the syllable dictionary, which is applied to get syllable sequences using the CTC prefix beam search with beam size 10. Syllable sequences are converted into character sequences by the decoder block using greedy search. The final decoding results are sorted by the weighted sum of the CTC score and the attention score. This process is conducted in a non-autoregressive way.

\subsection{Main Results}
The performance of the models was
evaluated based on character error rate (CER) without using language models, and we show the syllable unit error rate (SUER) calculated by the syllable sequence for reference. we conducted experiments to compare end-to-end ASR models using different modeling units. The results of character-based Transformer and Conformer models are also obtained by the two-pass decoding method \cite{yao2021wenet}.

As shown in Table~\ref{tab:transformer}, we compare the multi-level modeling units method with character-only modeling units based Transformer \cite{yao2021wenet} and RNN-T \cite{wang2021cascade} networks. We can see that our proposed method achieves 4.6\%/5.2\% character error rate on the AISHELL-1 dev/test sets, which surpasses the character-based RNN-T \cite{wang2021cascade} and Transformer \cite{yao2021wenet} networks. Furthermore, we compared our method with the models using syllable modeling units, and the proposed method outperforms the cascade RNN-T \cite{wang2021cascade} and decoupling models \cite{yuan2021decoupling} as well. 

The proposed method also achieves good results with the Conformer network. As shown in Table~\ref{tab:conformer}, when applied to the Conformer architecture, our method achieves CER of 4.1\%/4.6\% on the dev/test sets. It outperforms the character-based Conformer model as well as recently emerged methods using the Conformer backbone \cite{guo2021recent,tian2021consistent}. 

\subsection{Ablation Study}
We study the effects of InterCE loss for the final performance using the Conformer network. First, we study adding character-level InterCE loss to character-only based conformer network, and see a slight improvement in the dev set. Second, we try removing InterCE loss from our multi-level modeling units framework and find that it degrades the results by 0.1 on dev/test sets.
These results in Table~\ref{tab:ablation} suggest the advantage of unifying syllable-level modeling units and character-level modeling units in an end-to-end model. And the InterCE loss makes additional improvements under this multi-level modeling units framework.

\begin{table}[t]
  \caption{Ablation study of InterCE loss. 
  }
  \label{tab:ablation}
  \centering
  \begin{tabular}{lcc}
   \hline
    Model Architecture& Dev set & Test set  \\
   \hline
    Character-based Conformer & 4.4&4.7  \\
    \hspace{0.2cm}+ character-level InterCE loss &4.3&4.7 \\
   \hline
    Multi-level modeling units (I4) & 4.1 &4.6  \\
    \hspace{0.2cm}- character-level InterCE loss &4.2&4.7 \\
   \hline
  \end{tabular}
\end{table}

\section{Conclusions}
In this paper, we propose a novel end-to-end mandarin speech recognition method equipped with multi-level modeling units, which allows the model to integrate multi-level information. During the inference process, the input feature sequences are converted into syllable sequences and subsequently converted into Chinese characters within a unified end-to-end model. We further utilize the \textit{InterCE} loss to improve the accuracy of converting syllables to Chinese characters. The proposed model achieves competitive results and outperforms recently published literature on the AISHELL-1 benchmark.

\bibliographystyle{IEEEtran}

\newpage
\bibliography{mybib}

\end{document}